\documentclass[conference]{IEEEtran}
\usepackage{cite}
\usepackage{graphicx}
\usepackage[cmex10]{amsmath}
\usepackage{array}
\usepackage{mdwmath}
\usepackage{mdwtab}

\usepackage{boxedminipage} 
\usepackage{url}

\begin{document}

\title{Siamese Networks for Semantic Pattern Similarity}

\author{

\IEEEauthorblockN{Yassine Benajiba\IEEEauthorrefmark{1},
Jin Sun\IEEEauthorrefmark{2}, 
Yong Zhang\IEEEauthorrefmark{2},
Longquan Jiang\IEEEauthorrefmark{2},
Zhiliang Weng\IEEEauthorrefmark{2} and
Or Biran\IEEEauthorrefmark{3}}
\IEEEauthorblockA{
\IEEEauthorrefmark{1}Symanto,
\IEEEauthorrefmark{2}Mainiway,
\IEEEauthorrefmark{3}Elemental Cognition
\\
\IEEEauthorrefmark{1}\tt{yassine.benajiba@symanto.net},
\IEEEauthorrefmark{3}\tt{orb@elementalcognition.com} 
\\
\IEEEauthorrefmark{2}\tt{\{jin.sun,yong.zhang,longquan.jiang,zhiliang.weng\}@mainiway.com}
}

}

\maketitle
% --------------------------------------------
\begin{abstract}
Semantic Pattern Similarity is an interesting, though not often encountered NLP task where two sentences are compared not by their specific meaning, but by their more abstract semantic pattern (e.g., preposition or frame). We utilize Siamese Networks to model this task, and show its usefulness in determining SQL patterns for unseen questions in a database-backed question answering scenario. Our approach achieves high accuracy and contains a built-in proxy for confidence, which can be used to keep precision arbitrarily high.
\end{abstract}
% --------------------------------------------

\IEEEpeerreviewmaketitle

\section{Introduction}\label{sec:intro}
The {\em Semantic Textual Similarity} (STS) task \cite{agirre_et_al_2012}, which places pairs of sentences on a scale ranging from completely unrelated to perfect paraphrases (with intermediates such as topic similarity, partial coverage and near-identity), has gained much attention in recent years. In addition to being interesting on its own as a step towards semantic modeling of sentences, STS is an important (explicit or implicit) sub-task for many NLP applications such as Machine Translation, Generation, Summarization and Question Answering. Sentence-level similarity is significantly harder than word- or term-level similarity because of the variant length and semantic complexity of sentences, which makes it harder to model with traditional approaches (e.g., bag-of-words, tf-idf and similar models). Approaches to STS include pooling of word-level embeddings \cite{mikolov_et_al_2013}, fixed-length encodings of sentences \cite{le_mikolov_2014,kiros_et_al_2015,tai_et_al_2015}, matrix factorization approaches \cite{guo_diab_2012,zhao_et_al_2014} and neural similarity models \cite{he_et_al_2015,mueller_thyagarajan_2016}.

We explore the similar but less frequently encountered task of {\em Semantic Pattern Similarity} (SPS). Similar to STS, this task brings about its own challenges as a semantic task and is potentially useful for NLP applications. Semantic patterns are conceptually templates in which certain types of arguments are expected, and which model some abstraction of the semantics of a text unit - often an abstraction of entities, attributes and relations. One common type of semantic patterns includes linguistic semantic representations such as prepositions, frames and AMR \cite{banarescu_et_al_2013}, usually encountered in the context of the Semantic Role Labeling or Semantic Parsing tasks; however, semantic patterns may be formed with other structured representations of text, which result in a different kind of abstraction. 

\renewcommand*{\thefootnote}{\fnsymbol{footnote}}

\footnotetext[3]{Work done while at Mainiway AI Lab}

\renewcommand*{\thefootnote}{\arabic{footnote}}

Like STS, in SPS we are given two sentences. Instead of modeling the similarity of the sentences, however, we want to model only the similarity of the underlying semantic patterns. For example, the sentences ``John sold a car to Mary in 2016'' and ``Bob bought an apple from Jane on Sunday'' have identical semantic patterns. The model must therefore learn to distinguish between text relevant to the pattern and text relevant only to the specific sentence.

In this paper, we use SQL queries as proxies for semantic patterns. We use the recently introduced WikiSQL data set \cite{zhong_et_al_2017}, which contains over $87,000$ natural language questions aligned with SQL queries that produce the answer (from a known database). The benefits of using this data set are threefold: first, it is unusually large for this sort of annotated aligned data set; second, it is relatively straightforward to define a non-binary distance metric between SQL templates; and third, succeeding in the SPS task for this data set represents a large step towards the important real-world application of SQL-backed question answering. To make this third point more concrete, we evaluate our SPS approach by attempting to find, for an unseen single question, a question in the training data which shares a SQL template with it. In other words, we use our SPS model for the more challenging task of finding the semantic pattern (SQL template, in this case) of a new question. As an example of the usefulness of this task, it can be used as a sub-task in a SQL-backed QA scenario: first find the SQL template for a question; then fill in the blanks using named entities from the question or synonyms thereof, a task analogous to semantic parsing. 

To model the similarity of two questions, we use recurrent Siamese neural networks, an architecture that was successfully used for the closely related STS task \cite{mueller_thyagarajan_2016}. Our model achieves a high accuracy, and we show that the predicted similarity acts as a confidence score, so that thresholding it can keep precision arbitrarily high at the cost of not handling all cases.

\section{Related Work}\label{sec:rw}
While SPS, to the best of our knowledge, has not explicitly been pursued as a task, it is alluded to in the literature of related tasks. In addition to work on the STS task, which is briefly surveyed in the previous section, some relevent work exists in the context of paraphrasing. 
%newcite
\cite{biran_et_al_2016} mine {\em paraphrasal templates} - groups of concrete textual templates which would be paraphrases if filled with the same entities - from Wikipedia. Their approach relies on first finding and removing entities, and then clustering the remaining templates in a lexical vector space. In contrast, we model sentences directly in semantic pattern space. Earlier examples from the paraphrasing literature include 
%newcite
\cite{sekine_2005}, which uses heuristic rules to find short templated paraphrases, and 
%newcite
\cite{barzilay_lee_2003} who produce {\em slotted lattices} from a comparable corpus which contains paraphrases. 

In Natural Language Generation, 
%newcite
\cite{angeli_et_al_2010} used similar grouped templates, but did not use a similarity metric (instead relying only on entity types to group templates). 
%newcite
\cite{duma_klein_2013} extract semantic templates from Wikipedia pages by aligning them with entities from Semantic Web data. 
%newcite
\cite{bowman_et_al_2016} and 
%newcite
\cite{guu_et_al_2017} generate novel sentences by performing edits on another sentence, which tend to preserve the original semantic pattern. In both cases, the semantic pattern similarity is not modeled directly, but emerges from the model's learned behavior of making certain types of small lexical edits. 

While not as directly related, many standard practices in NLP apply some form of abstract text matching. For example, in Machine Translation, {\em alignment templates} \cite{och_ney_2004} can be thought of as a kind of semantic pattern matching across languages, while in Information Extraction, Hearst Patterns \cite{hearst_1992} and similar techniques where a set of lexicalizations of a single relationship are used to mine pairs of words or entities for which that relationship holds can be said to apply semantic pattern matching to these pairs. None of these approaches provide a semantic similarity score for arbitrary sentences. 

\section{Data}\label{sec:data}
We use the WikiSQL data set \cite{zhong_et_al_2017}, which contains aligned pairs of questions and SQL queries over database tables collected from Wikipedia. As described in the paper, the questions and queries are created with a hybrid algorithmic/templated approach and human editors (on Amazon Mechanical Turk) who perform the final matching and filtering. 

WikiSQL is the largest aligned text-SQL dataset publicly available. The queries are limited to having one select column, no nested queries, and no joins, which may limit its usefulness for real-world applications, but is an attractive property for research purposes as it constrains the problem domain. 

The data is fairly noisy: some questions are incoherent, and some do not match their paired SQL query. The column types of the database tables are not always correct, and most are mistakenly labeled as ``text'' by default. To alleviate these problems, we did some cleaning of the data. First, we used regular expressions to find numeric and date columns and correctly assign their types; and second, we removed questions that were too short or contained few alphabetic characters. We will release the resulting adjusted data set with a detailed description of the adjustments upon acceptance.

\section{Our Approach}\label{sec:our_approach}
Given an unseen question, our goal is to find, in a pool of questions, another question with the same semantic pattern (SQL template). 

We employ a Siamese LSTM regression model to predict the similarity of the SQL templates of two questions (Section \ref{subsec:siamese}). Instead of comparing the unseen question to the entire training set, which would be costly, we cluster the training set ahead of time using a lexical representation of the questions (Section \ref{subsec:lexic_clustering}), and compare a new question only to the members of its nearest cluster.

% --------------------------------------------
\subsection{SQL Structure Distance}\label{subsec:sql_dist}
We define the SQL structure distance as:
\begin{displaymath}%TODO write equation with binary functions for comparing type, etc and then for binary function make a separate equation to describe it
SQLSD(sql_i, sql_j) = \sum_{c \in C} I(sql_i(c),sql_j(c))
\end{displaymath}
\noindent where $I$ is an indicator function that equals $1$ when its two arguments match. $sql_i(c)$ is a function that indicates the value of the SQL query $sql_i$ for the SQL constituent $c$. Finally, $C$ is the set of constituents we take into consideration to compute the distance between two queries. It consists of the {\em type} of the SELECT column (text, number, date); the {\em aggregator} of the selection (none, COUNT, SUM, etc.); and the number of elements in the WHERE clause for each condition ($=$, $>$, $<$, etc.). These are enough because of the limited complexity of the data set, as described in Section~\ref{sec:data}. 

% --------------------------------------------
\subsection{Siamese LSTM Regressor}\label{subsec:siamese}
Siamese networks, designed as a way to compare two objects, were first introduced by 
%newcite
\cite{bromley_et_al_1994}. 
By focusing on comparison, the network is not tied to a particular set of classes.
A Siamese network can be diagrammed as two branches (although both branches share the same parameters), both feeding to one layer that performs some distance measure to get an output. The idea behind this architecture is to learn a transformation function that transforms the input representation into a space where similar objects have similar representations.

\begin{figure}
\centering
\includegraphics[width=0.45\textwidth]{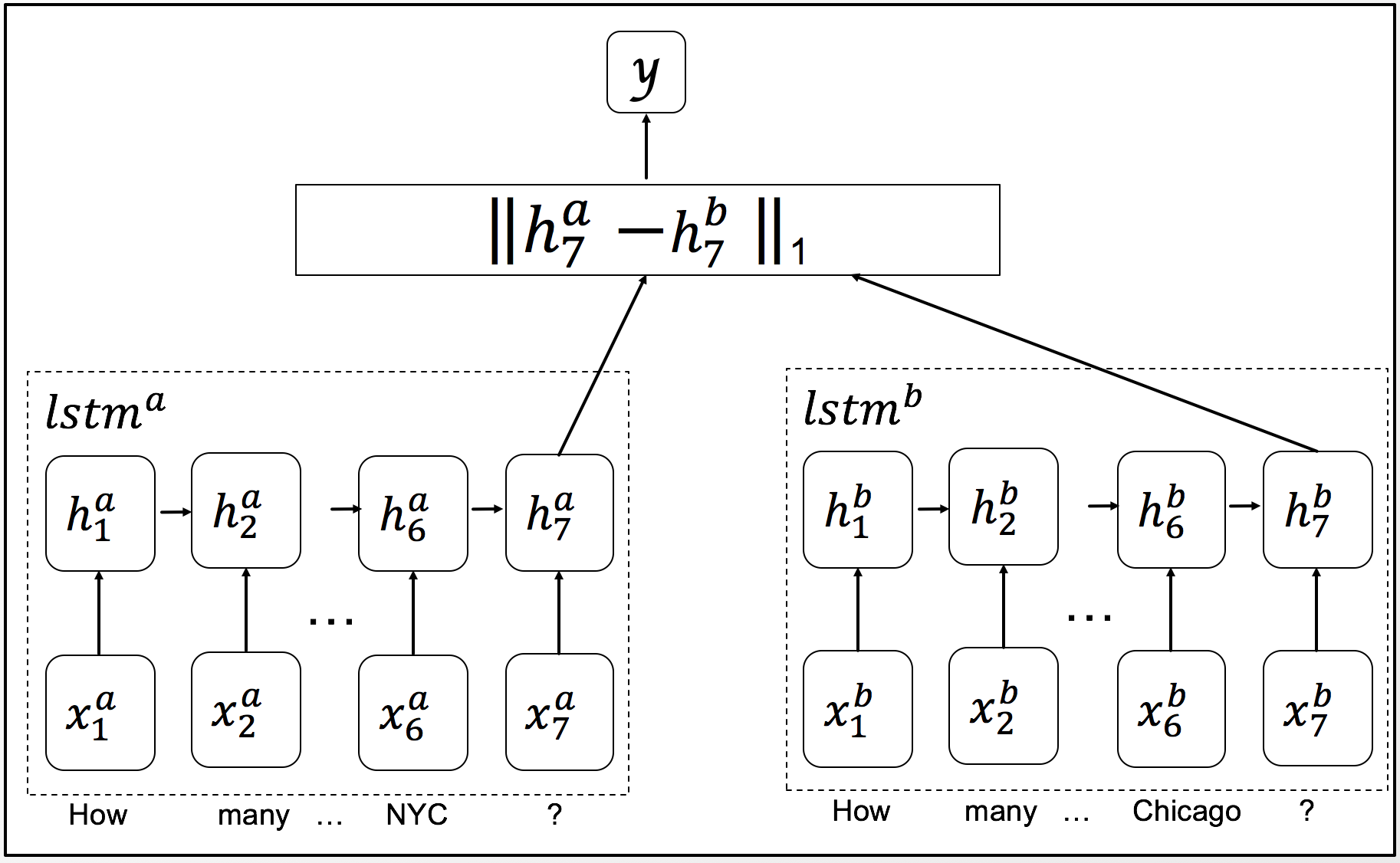}
\caption{\label{fig:arch}Siamese LSTM Architecture}
\end{figure}

%newcite
\cite{mueller_thyagarajan_2016} use Siamese networks to compare pairs of sentences. Unlike %RUSLANPAPER
%newcite
\cite{bromley_et_al_1994}, they use a recurrent network to learn the transformation function as it is more appropriate for modeling language. We use a similar architecture, shown in Figure \ref{fig:arch}. 
The major difference is that given the nature of our distance metric, we train our network to learn a regression function. The loss function is the mean squared error from the distance measure $SQLSD$.
%, defined in Section~\ref{subsec:sql_dist}.

\subsection{Question Lexical Clustering}\label{subsec:lexic_clustering}
Comparing an unseen question to every question in the training set would be prohibitively (and unnecessarily) costly for a real-world QA scenario. To reduce the size of our search space, we first cluster the training set instances using a one-hot lexical representation of the questions, using as a vocabulary all words appearing in the training set with a frequency $> \alpha$ (in our experiments, we set $\alpha = 50$. This number was found empirically by tuning on the dev set). We then use k-means to cluster the data in this vector space. With $k=500$, we obtain an average cluster size of $122.6$.

The one-hot representation is most adequate for the problem at hand because the questions are typically short, and the frequency of each term is almost always $1$. In addition, because we are interested in the semantic pattern rather than the semantics themselves, functional words are important; word embeddings or other distributional representations may not discriminate between them. 

Because of the choice of $\alpha$, the questions are clustered mostly by these functional words; as a result, questions in different clusters tend to be questions with different functional words (e.g., ``how'' and ``many'' vs. ``who'' and ``was''). The clustering therefore maximizes the inter-cluster SQL structure distance: while two questions in the same cluster are not guaranteed to have the same SQL template, questions in different clusters are much {\em less} likely to share a template, and therefore it is relatively safe to ignore them and focus only on the most similar cluster at prediction time.

% --------------------------------------------
\section{Experiments and Results}\label{sec:expts_rslts}
% --------------------------------------------
In our experimental set up, for each question in the test (or dev) set, we try to find a question in the training set which has the exact same SQL template, and evaluate our success using the (binary) accuracy of the selection. 

One advantage of using a soft regression score for a binary task is being able to use a minimum threshold. In a real-world application, it is often useful to keep a very high accuracy for accepted instances, even at the cost of rejecting some instances (``unable to handle'' is a better response than a wrong answer). To that end, we define a threshold $\beta$, such that if there is no training question in the cluster for which the network predicts a distance $< \beta$ from the test question, we simply reject the test question. This results in a ``safe classification'' as defined by \cite{wang_et_al_2017}.

In our evaluation, we therefore track two quantities for varying values of $\beta$ (from $0.1$ to $2.5$): the ratio of correctly matched questions to non-rejected questions, and the ratio of correctly matched questions to all test questions. We also track the total number of rejected questions and of incorrectly rejected questions (i.e. questions for which there really is a match in the cluster).

We train our model with a single hidden layer of size $100$, over $25$ epochs with a batch size of $1024$. The input is composed of pre-trained $word2vec$ embeddings with $300$ dimensions. %\footnote{https://code.google.com/archive/p/word2vec/}

We compare our approach with two baselines: {\em embeddings} uses the cosine similarity of the average embeddings of the questions as a similarity metric and chooses the nearest question; {\em accept-all} is the case when $\beta$ is infinite and no questions are rejected, in which case we always take the nearest question in the cluster. 

\begin{figure}
\centering
\includegraphics[width=0.5\textwidth]{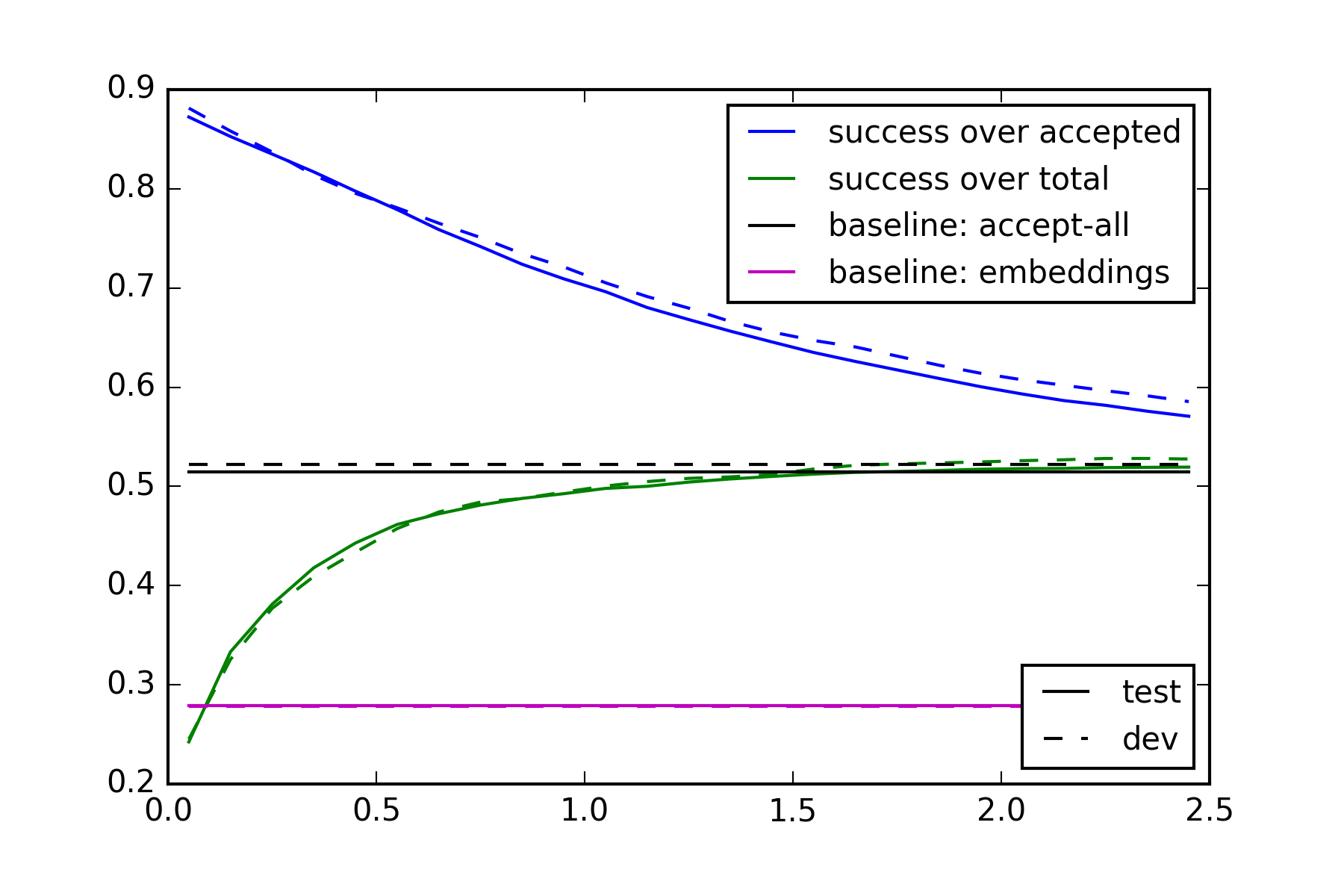}
\caption{\label{fig:accuracy}Our accuracy for different values of $\beta$ in comparison with our two baselines.}
\end{figure}
\begin{figure}
\centering
\includegraphics[width=0.5\textwidth]{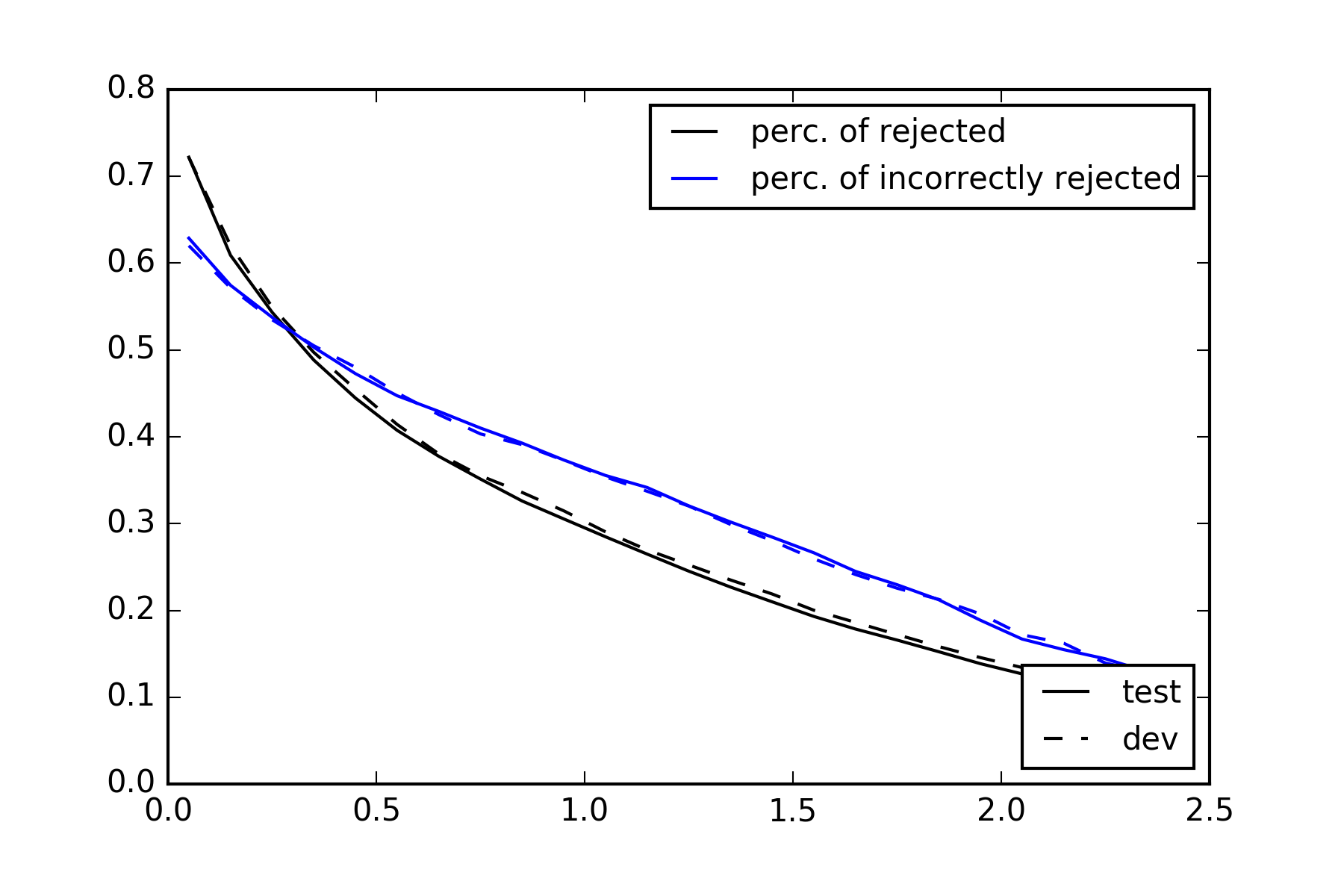}
\caption{\label{fig:rejected_prec}
\% of rejected questions for different values of $\beta$ and \% of times we incorrectly rejected a question.}
\end{figure}

As Figure \ref{fig:accuracy} shows, the results are almost identical for the dev and test sets.
%despite the difference in size (see Section \ref{sec:data}). 
We obtain an accuracy of $75\%$ with $\beta = 0.75$, at the price of rejecting $35\%$ of the questions, $60\%$ of them correctly rejected: it would be impossible to find the right answer in that cluster. A solution without a threshold would force the model to make a choice even when the test instance is very different from the what the model has been trained on; the confidence scores of neural classification models become highly unpredictable for instances that are sufficiently different from the training data \cite{wang_et_al_2017}.

The dramatic improvement in accuracy over the {\em embeddings} baseline, which achieves an accuracy of only $27.8\%$, shows that our model's representation is very different from the sentence's lexical embeddings, and it does seem to constitute a sort of ``semantic pattern space'', as we hypothesize. At the same time, using our model without using a threshold - as the {\em accept-all} baseline does - yields an accuracy of only $51.5\%$. Overall, the experiments suggest that the best solution is to combine the ability of recurrent Siamese networks to accurately transform the questions to a semantic pattern space with a threshold to classify selectively.

Figure~\ref{fig-examples} shows four example questions that share a SQL template. Despite the lexical and semantic differences, our model correctly identifies the structural similarity and matches them; the embedding baseline does not. 

%To illustrate the value of our approach, the following two questions share a SQL template: ``who is the player who played for Miami Sol and went to school at North Carolina State?'' and ``what is the lightweight value with no information model and the flexible value is unknown?''. Despite the lexical and semantic differences, our model correctly identifies the structural similarity and matches them; the embedding baseline does not. 

\begin{figure}
\small
\begin{boxedminipage}{0.48\textwidth} 
 - Who is the player who played for Miami Sol and went to school at North Carolina State?
 
 - Which method resulting in a win against Ed Mahone?
 
 - Tell me the host for midwest thomas assembly center.
 
 - At which track was Frank Kimmel the Pole Winner of the Pennsylvania 200? 
\end{boxedminipage}
\caption{Four questions that share a SQL template.}
\label{fig-examples}
\end{figure}

%To illustrate, our model succeeds in identifying that the questions: {\bf \emph{Who is the player who played for Miami Sol and went to school at North Carolina State?}} and {\bf \emph{What is the lightweight value with no information model and the flexible value is unknown?}}; are structurally similar whereas the lexical model fails to do so because of the different ways in which the questions are formulated.

\subsection{Question Representation}\label{subsec:q_rep}
As a secondary, informal evaluation of our approach, we visually inspect the quality of the embeddings produced by the LSTM. These embeddings are obtained by reading the state of the last hidden layer after processing a question with the trained LSTM. Consequently, each question is represented as a vector of dimension $100$ (size of the hidden layer) and is projected to 2 dimensions using t-SNE. %\footnote{the t-SNE and the plotting was done using Google Embedding Projector: http://projector.tensorflow.org/}.

\begin{figure}
\centering
\includegraphics[width=0.48\textwidth]{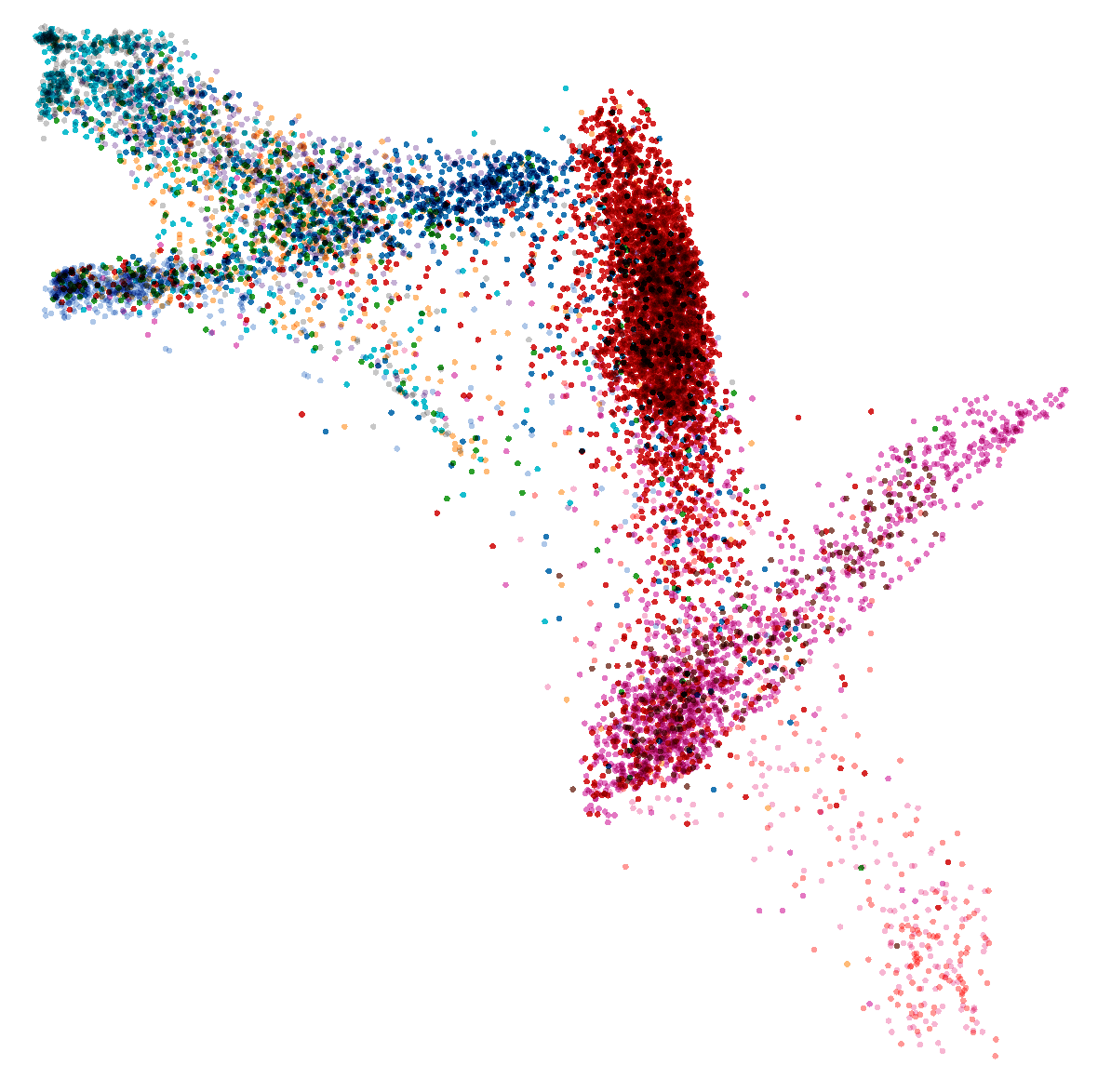}
\caption{\label{fig:q_rep}Siamese LSTM representation for the Most frequent types of SQL queries}
\end{figure}

Figure \ref{fig:q_rep} shows the projected embeddings. The colors of the dots represent the true class (SQL template) of the question. We exclude groups with less than 500 members since they are difficult to evaluate visually.

% --------------------------------------------
\section{Conclusion}
% --------------------------------------------
We show how to accurately and efficiently find the semantic structure of a sentence by comparing it with sentences with known structures, and evaluate our approach on questions aligned with SQL queries. Our results indicate that a combination of recurrent Siamese networks and nearest neighbor threshold validation yields high accuracy results. 

%While we obtain good results on the WikiSQL dataset, we would like to explore the effectiveness of this approach on additional domains and more general data . To extend this work in the future, we plan to experiment with text exhibiting other semantic structures, in particular AMR and semantic frames.

\bibliographystyle{IEEEtran}
\bibliography{main}

\end{document}